**Empirical Comparison of Encoder-Based Language Models and Feature-Based Supervised Machine Learning Approaches to Automated Scoring of Long Essays**

Kuo Wang[1], Haowei Hua[2], Pengfei Yan[3], Hong Jiao[3], Dan Song[4]

[1]Southern Methodist University, Dallas, TX, USA

[2]Princeton University, Princeton, NJ, USA

[3]University of Maryland, College Park, MD, USA

[4]University of Iowa, Iowa City, IA, USA

**Abstract**

Long context may impose challenges for encoder-only language models in text processing, specifically for automated scoring of essays. This study trained several commonly used encoder-based language models for automated scoring of long essays. The performance of these trained models was evaluated and compared with the ensemble models built upon the base language models with a token limit of 512?. The experimented models include BERT-based models (BERT, RoBERTa, DistilBERT, and DeBERTa), ensemble models integrating embeddings from multiple encoder models, and ensemble models of feature-based supervised machine learning models, including Gradient-Boosted Decision Trees, eXtreme Gradient Boosting, and Light Gradient Boosting Machine. We trained, validated, and tested each model on a dataset of 17,307 essays, with an 80%/10%/10% split, and evaluated model performance using Quadratic Weighted Kappa. This study revealed that an ensemble-of-embeddings model that combines multiple pre-trained language model representations with gradient-boosting classifier as the ensemble model significantly outperforms individual language models at scoring long essays.

**Keywords:** Automated Essay Scoring, Long Essay, Encoder-based Language Models, Machine Learning, Writing Assessment

**Word Count:** 3,876 words

# Introduction

Automated Essay Scoring (AES) has become instrumental in meeting the increased demand for quick, accurate, and objective grading. Early AES systems were developed primarily to address the logistical challenges associated with large-scale writing assessment, particularly the time, cost, and consistency required for human scoring (Warschauer & Ware, 2006). Initial systems, such as Project Essay Grade (PEG), relied heavily on surface-level textual features, including essay length, syntactic patterns, and lexical counts, to approximate human ratings of writing quality. These features were weak representations of writing quality, which may explain the limited instructional use of the first AES systems. The second generation of AES systems applied natural language processing to evaluate the students' writing on a wider range of linguistic, discourse, and semantic features. According to Ramesh and Sanampudi (2022), the contemporary generation of AES systems uses deep learning methods to model the latent relationships among content, organization, language, and style, leading to more accurate and robust scores. More recent research in AES focuses on applying state-of-the-art language models , or pretrained language models, which encode contextual, semantic, and syntactic information directly from raw text. As summarized by Ramesh and Sanampudi (2022), embedding-based approaches have substantially improved scoring accuracy by reducing feature engineering and enabling models to generalize across prompts and writing styles.

As AES proves to be efficient and accurate to support scalable, consistent, and rapid scoring, enabling broader use of writing tasks in large-scale tests and classroom assessments (Hussein et al.; Ramesh and Sanampudi; Xu et al., 2024). Though encoder-based language models such as BERT, RoBERTa, and DeBERTa demonstrated increased accuracy in automated scoring, they are limited by the maximum token length, which is often 512. Research increasingly recognizes that long and extended essays pose specific modeling and computational challenges beyond standard short- or medium-length AES tasks. To investigate the performance of encoder-based language models in automated scoring long essays, this study investigates the impact of embedding source and ensembling strategies on the scoring performance of long student essays. Such settings also enable a direct comparison between single embedding models and the feature-based ensemble methods, while using the same evaluation criteria and data splits.

The remainder of this paper proceeds as follows. First, we provide a concise literature review of related work on automated essay scoring (AES), specifically methods based on deep learning, language models, and hybrid ensemble methods. The methods section explains the dataset, approaches to preprocessing long essays, and model architectures ranging from single-model baselines to embedding-level ensemble methods and feature-based ensemble classifiers. We provide model comparisons using the Quadratic Weighted Kappa (QWK; Fleiss & Cohen, 1973) and conclude the paper with discussion of limitations and future directions.

## Encoder-Based Language Models

Language models range in size from a few million to hundreds of billions or even trillions of parameters. In the current study, we follow convention and define small language models (SLMs) and large language models (LLMs) as relative categories, not defined by hard cutoffs in size. This is done for ease of comparison and to enable a more systematic evaluation of model design choices in AES.

BERT and its variants are considered as small language models (SLMs), including BERT (Devlin et al., 2019), RoBERTa (Liu et al., 2019b), DistilBERT (Sanh et al., 2019), and DeBERTa (He et al., 2021). BERT-base and BERT-large, for instance, have 110 and 340 million parameters, respectively. RoBERTa-base and RoBERTa-large have 125 and 355 million parameters, respectively, and DeBERTa-base and DeBERTa-large have 140 and 400 million parameters, respectively. DistilBERT, a distilled variant of BERT, has 66 million parameters, about 60% of BERT-base (Sanh et al., 2020). Large language models (LLMs), on the other hand, generally have a much larger parameter capacity and are typically trained with more open-ended generative or instruction-following use cases in mind. For instance, Gemma-2B has about 2 billion parameters (Riviere et al., 2024), T5 has a large variety of different sizes, with the larger variants like T5-11B having about 11 billion parameters (Raffel et al., 2023), and LLaMA-3.2-3B has about 3 billion parameters (Grattafiori et al., 2024). In this study, models of this size and above are grouped as LLMs to differentiate them from encoder-based SLMs, based on their architectural, training, and usage differences rather than their absolute number of parameters.

The contextualized word embeddings produced by these SLMs can be readily used for AES. As shown in Figure 1, the BERT-based word embeddings can serve as input to a downstream classification model that assigns a score to the essay. Embedding representations from the larger BERT variants can have dimensions up to 512 × 1,024. Since most BERT-based models are limited to a maximum input length of 512 tokens, they cannot accommodate essays that are longer than this length, which is not a rare issue in essay writing.

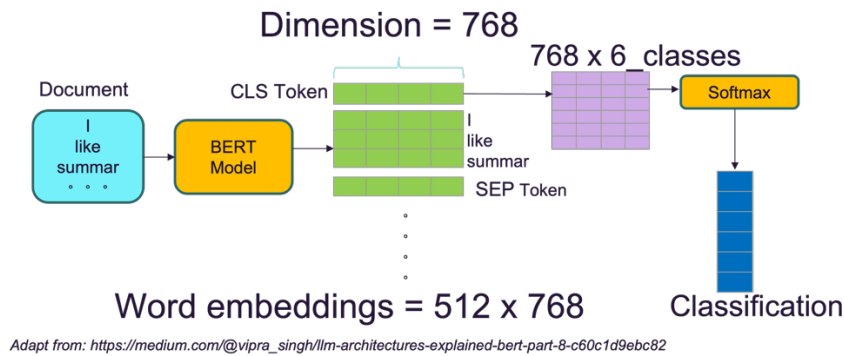

*Figure 1*

*Use Word Embeddings for AES*

**Automated Essay Scoring**
Previous AES models were highly dependent on hand-crafted linguistic features such as syntax, grammar, and essay length. ETS's e-rater and Vantage Learning's Intellimetric represent examples of such an approach, offering feedback primarily based on superficial textual features (Warschauer & Ware, 2006). Alongside raw text embedding, feature engineering is essential for automated scoring, particularly when incorporating linguistic features (Bulut et al., 2024). Features that can be elicited from essays, including syntactic complexity, lexical richness, discourse markers, argument structure, and semantic coherence, have been shown to improve scoring performance (Hussein et al., 2019; J. Liu et al., 2019; Shang et al., 2023; Uto, 2021). These features are often combined with machine learning methods such as ensemble learning and boosting, which aggregate the predictive capacity of multiple models to improve the generalizability and robustness of AES (Chen & Guestrin, 2016; Zhang, 2024). Ensemble methods can couple predictions of deep neural networks and conventional classifiers while boosting methods like Gradient-Boosted Decision Trees (GBDTs; Shi et al., 2022), Light Gradient Boosting Machine (LightGBM; Ke et al., 2017), and eXtreme Gradient Boosting (XGBoost; Chen & Guestrin, 2016) tend to learn well from feature interactions and residual errors.

The development of deep neural networks (DNN) marked the end of manual feature engineering. The use of Convolutional Neural Networks (CNNs), Long Short-Term Memory (LSTM) networks, and transformer models like BERT became widespread, significantly enhancing scoring precision through their inherent ability to learn complex textual features automatically (Shang et al., 2023; Uto, 2021). For example, Shang et al. (2023) combined LSTM semantic models and XGBoost, to demonstrate the value of hybrid models for analytic scoring of essay in dimensions of writing quality and content relevance.

Recent studies have demonstrated that AES systems employ transformer-based models, such as BERT and its variants, leading to significant improvements over classical neural networks. These models learn from extensive language corpora and remain highly effective in understanding subtle content and context in an essay (Shang et al., 2023; Wang et al, 2022???). These encoder models built on transformer architectures with the multi-head attention mechanism (Vaswani et al., 2017) have become foundational in natural language processing (NLP) tasks and AES (Clark et al., 2019; Devlin et al., 2019; Fernandez et al., 2023; Lotridge, 2023; Reimers & Gurevych, 2019; Shermis & Wilson, 2024). Transformers excel at handling the contextual dependencies of word sequences, making them well-suited for evaluating linguistic features such as coherence, cohesion, and the argument structure of students' essays. However, despite their effectiveness, most SLMs encounter technical limitations when processing long essays that exceed typical input token limits (e.g., 512 tokens) due to the maximum token limit imposed during input tokenization.

While transformer-based AES are typically limited to 512 tokens, which is insufficient for many long essays (Hua et al., 2025; Ormerod et al., 2021), automated scoring of extended essays (e.g., ICLE with ~680–1,090 subword tokens; Learning Agency Lab AES 2.0 with up to 1,656 words) is often subject to issues of truncation, input length variability, and bias toward shorter texts (Hua et al., 2025???; Mim et al., 2021). There is clear movement toward handling long essays via hierarchical encoders, length-aware transformers, and LLM-based summarization (Hua et al, 2025), but fully robust, validated AES for truly long, authentic writing remains an active research area.

Approaches dealing with long-document modeling can be categorized into three main types: long-sequence transformers, chunk-and-aggregate systems, and ensemble-of-embeddings approaches. The first centers on long-sequence transformers, extending attention to process thousands of tokens directly (e.g., Longformer, BigBird, Clinical-Longformer, LED) and showing impressive results in clinical (Li et al., 2022) and financial contexts (Khanna et al., 2022), and on large-scale benchmark test suites (Boytsov et al., 2024). Although such models retain global context, they are computationally expensive and often yield diminishing returns on tasks where relevance can be captured by a small number of passages. The second category relies on chunk-and-aggregate methods, in which documents are divided into chunks of tractable size and subsequently re-aggregated using heuristic or learned aggregation rules (e.g., MaxP, Birch, PARADE). In particular, PARADE (Li et al., 2023) has recently reported that hierarchical passage-level aggregation of passage representations can compete with or surpass full-sized LLMs, such as T5-3B, while being more efficient. However, it exploits its advantages best when there are rich signals across multiple passages, but it is less effective in domains where a single passage dominates. In contrast to these two approaches, our work introduces a third pathway: using an ensemble model of embeddings to combine aggregates of different pretrained LLM encoders' embeddings using gradient-boosting learners (e.g., XGBoost, LightGBM), including DeBERTa, Longformer, and BigBird. This approach avoids the computational overhead of the long-sequence model and the architectural overhead of passage-level aggregation, instead leveraging the complementarity of cross-encoders.

Typically, transformer-based models like BERT and its variations (Clark et al., 2020; Devlin et al., 2019; He et al., 2021; Y. Liu et al., 2019; Sasaki & Masada, 2022) allow for input sequences of up to 512 tokens, thus limiting their ability to directly handle longer pieces of text. To address this issue, researchers introduced specialized transformer model versions like BigBird, Hierarchical BERT (Kong et al., 2022; Zaheer et al., 2021; Zhang et al., 2019), and Longformer (Beltagy et al., 2020), which were specifically designed for sparse attention or hierarchical encoding for the handling of longer sequences of tokens. BigBird supports 4,096 tokens.

Hierarchical BERT processes long documents by breaking them into smaller segments (which can be the standard 512-token BERT size). Like BigBird, Longformer can also process up to 4,096 tokens by adding local windowed attention with global attention mechanisms. In contrast, the Longformer Encoder-Decoder (LED) can process up to 16,384 encoder input tokens (AI2, 2020/2024; Pang et al., 2022).

More recently, the advent of generative language models (GLMs), such as GPT-5 (OpenAI, 2026), Claude (Anthropic, 2026), and Google Gemini (Anil et al., 2024), has significantly expanded the capacity to process long documents due to their much larger context windows. GPT-5 has a maximum input size of approximately 400,000 tokens; Claude sonnet 4.5 models up to 200,000 tokens; and Gemini 3 pro has an input context window of 1 million tokens—sufficient for full-document analysis without cutting. These larger capacities make GLMs particularly valuable for AES tasks, where whole-document analysis is vital. However, data privacy and security are always major concerns when using GLMs in educational assessment applications, including essay scoring.

Hybrid solutions are also emerging. These can either be models that fuse traditional feature-based approaches with a large language model, or models that utilize the outputs of multiple (Chebrolu et al., 2005; Ding et al., 2022, 2024; Kag et al., 2022). These architectures typically use LLMs to produce a dense representation of the essay, to which handcrafted or automatically selected linguistic features are appended as additional input to the final classifier. Yao (2024) uses hard voting across five independently trained models, each employing different architectural and training strategies. All models use DeBERTa-v3-large as the backbone language model, which provides rich contextual representations through its disentangled attention mechanism and large-scale masked language modeling (MLM) pretraining. This backbone is either frozen or fine-tuned depending on the design of each experiment, serving as the foundation for downstream prediction heads. Additionally, handcrafted text features, including paragraph-level features such as length, spelling errors, word/sentence counts, sentence-level features such as length and word counts, and word-level features (word lengths, distributions), are integrated into the entire AES system to better capture semantic and linguistic features, thereby boosting overall system performance. All models adopt a two-stage training procedure: pretraining on the larger Kaggle-Persuade subset to acquire general knowledge, followed by fine-tuning on the Kaggle-Only subset to adapt to the target distribution, which is closer to the hidden test set. This staged approach mitigates distributional bias caused by dataset imbalance and improves generalization. Several models incorporate ordinal regression objectives to better capture the ordered nature of essay scores, while others experiment with PET-like training and hierarchical pooling strategies to enrich representation learning. To further refine predictions, threshold optimization using an adapted Nelder–Mead search algorithm and a custom for-loop to fine-tune the last threshold is implemented, boosting cross-validation (CV) scores. Finally, individual model predictions are aggregated via a hard voting ensemble, resulting in improved stability and performance across validation metrics.

## Purposes of the Study

This study aims to develop automated scoring models for long essays and compare the performance of different models, including encoder-based SLMs, their ensembles, and feature-based ensemble models. The research question is proposed as how these various models compare in the automated scoring of long essays.
1. How do encoder-based SLMs perform in scoring long essays with truncation?
2. Can ensemble of encoder-based language models improve scoring accuracy?
3. Can ensemble models based on features extracted from the full essay text without truncation improve the performance for the automated scoring of long essays?

This study contributes to automated scoring in two aspects. First, this study conducted a comprehensive empirical comparison to show the state-of-the-art encoder-based language models are not sufficient for automated scoring of long essays. Second, we present an ensemble-of-embeddings framework that combines multiple pre-trained language model representations using gradient boosting classifiers (XGBoost, LightGBM). This solution offers a computationally efficient and cost-effective alternative for scoring extremely long essays.

## Methods

**Data**

This study utilized the Learning Agency Lab—Automated Essay Scoring 2.0 dataset, which comprises 24,000 argumentative essays, including 17,307 in the training set (Kaggle.com, 2024). Essay scores range from 1 to 6, with higher scores indicating better performance. The distribution of scores is imbalanced, with score 3 dominating (6,280 samples) and only 156 samples receiving a score of 6 (Figure 2). Essay lengths range from 150 to 1,656 words. Specifically, 2,969 essays contain more than 500 words, 2,659 essays are between 512 and 1,024 words, and 30 essays exceed 1,024 words (see Figure 3 for an overall distribution plot on a log scale).

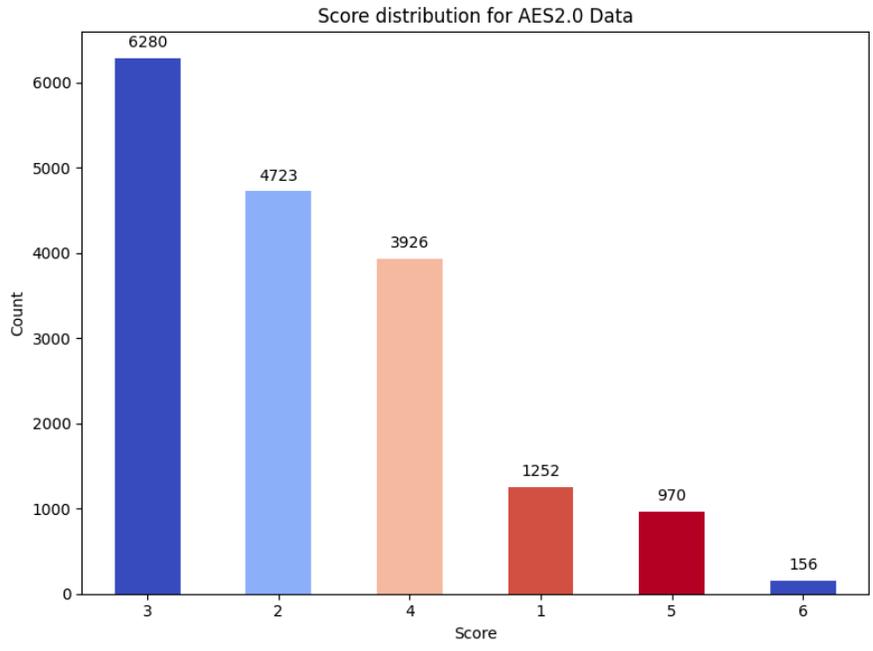

Figure 2
*Score Distribution for AES 2.0 Data*

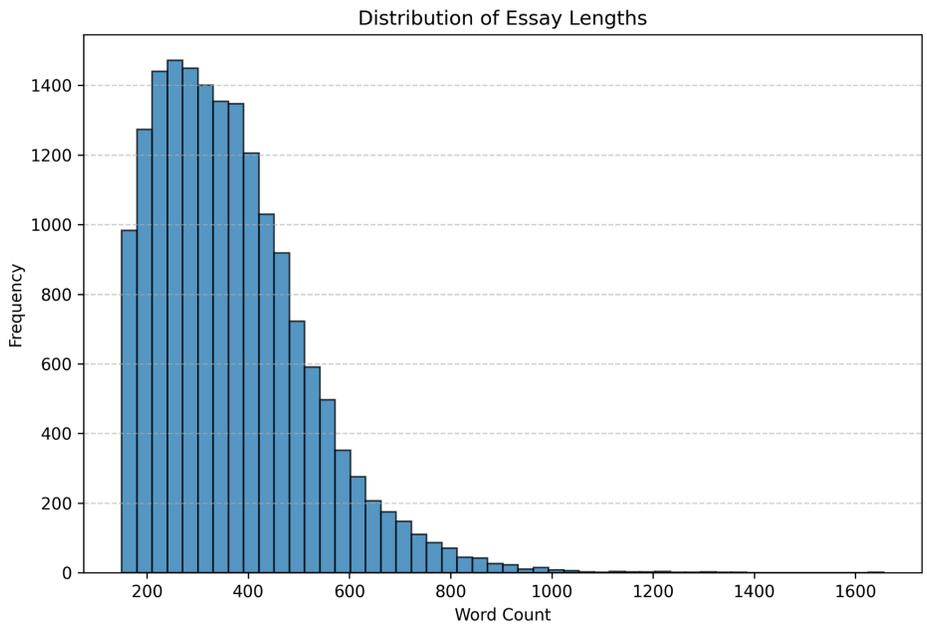

Figure 3
*Essay Length Distribution for AES 2.0 Data*

**Research Design**

Most existing long-text modeling work has utilized long-sequence transformers, increasing context windows or chunk-and-aggregate techniques, dividing and aggregating passages, but our work pursues a third option: an ensemble-of-embeddings framework utilizing a variety of pre-trained language models, gradient boosting classifiers (e.g., XGBoost, LightGBM), and feature-based voting ensembles to have efficient and reliable scoring of long essays.

To answer the research question, we compared three sets of models. We first trained several encoder-based language models, including BERT, RoBERTa, DistilBERT, DeBERTa (small and large), DeBERTa v3 (large), Longformer (small), and BigBird-RoBERTa, as baseline. Secondly, we created ensemble models, where embeddings from multiple pre-trained SLMs were combined either by passing them as input to a Multilayer Perceptron (MLP), as the latter ones as a group, or where single embeddings (BERT, RoBERTa, DeBERTa) were combined with techniques like gradient boosting, such as XGBoost, LightGBM. Thirdly, we experimented with feature-based ensemble models, in which hand-designed linguistic features were combined with gradient-boosting classifiers. Model performance was measured based on the Quadratic Weighted Kappa (QWK), i.e., the ordinal nature of the scores on the essays being taken into account (Shermis, 2014, 2015). All implementations are conducted in Python 3.11.

**Data Preprocessing**
The data was split into three subsets: training, validation, and test. The final sample sizes for each subset are 13,845, 1,731, and 1,731, respectively. For feature-based approaches, we extracted text features that include 1) paragraph features (such as length, spelling errors, and word/sentence count); 2) sentence features (length, word count, etc.); 3) word-level features (word lengths, distributions); 4) text vectorization features, including TF-IDF vectorizer and count vectorizer.

**Baseline Models**
We trained multiple SLMs, including BERT, RoBERTa, DistilBERT, and DeBERTa as baseline models. Each model was trained with the same hyperparameter settings for easier comparison between models. The essays were tokenized to a maximum sequence length of 512 tokens, and each model was trained for three epochs with a batch size of 32. Training was performed using the Adam optimizer with an initial learning rate of $5 \times 10^{-6}$ and a cosine scheduler with a short warm-up period, followed by gradual decay. The classification head was trained with sigmoid activation and binary cross-entropy loss. Ordinal encoding was used to represent the ordinal relationship of the essay scores. Pipelines of cached and prefetched datasets were used for training and validation datasets for efficiency. The best model was selected based the highest QWK on the validation dataset through checkpointing. All experiments were run with a fixed random seed (42) to ensure reproducibility.

**Ensemble Models**

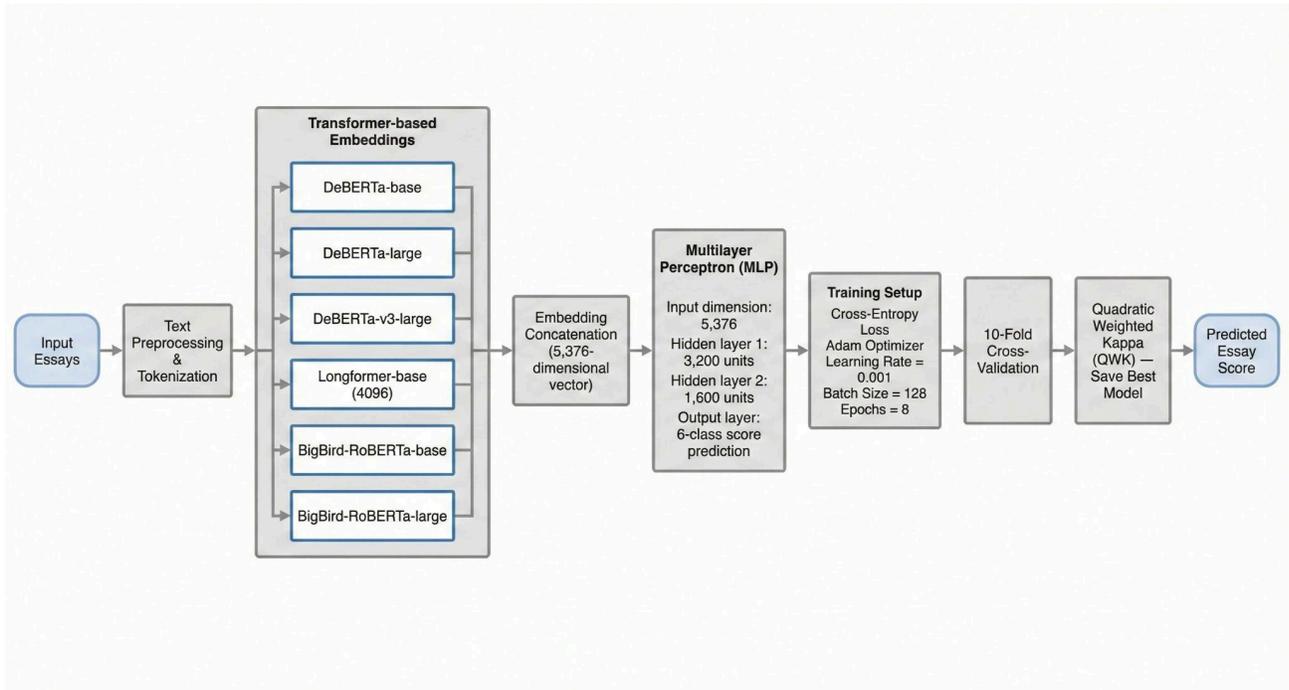

Figure 4: Ensemble Models Architecture

We explored two types of ensemble approaches. One is with a simple neural networking implementation (See Model 1 in Table 1). We concatenated the embeddings of multiple models, including DeBERTa base and large, DeBERTa v3 large, Longformer base-4096, and BigBird-RoBERTa base and large, as an input to a two-hidden-layer Multilayer Perceptron (MLP). The MLP model was configured with an input dimensionality of 5,376, two hidden layers of 3,200 and 1,600 units respectively, and an output layer of 6 classes. Training used the cross-entropy loss function with the Adam optimizer, with a learning rate of 0.001, a batch size of 128, and 8 epochs. The model is trained using 10-fold cross-validation and evaluated with QWK to save the best model. Another exploration is to combine a single BERT-based embedding, including BERT, DeBERTa, and RoBERTa, with the XGBoost and LightGBM approaches (See Models 2, 3, and 4 in Table 1).

**Feature-Based Models**

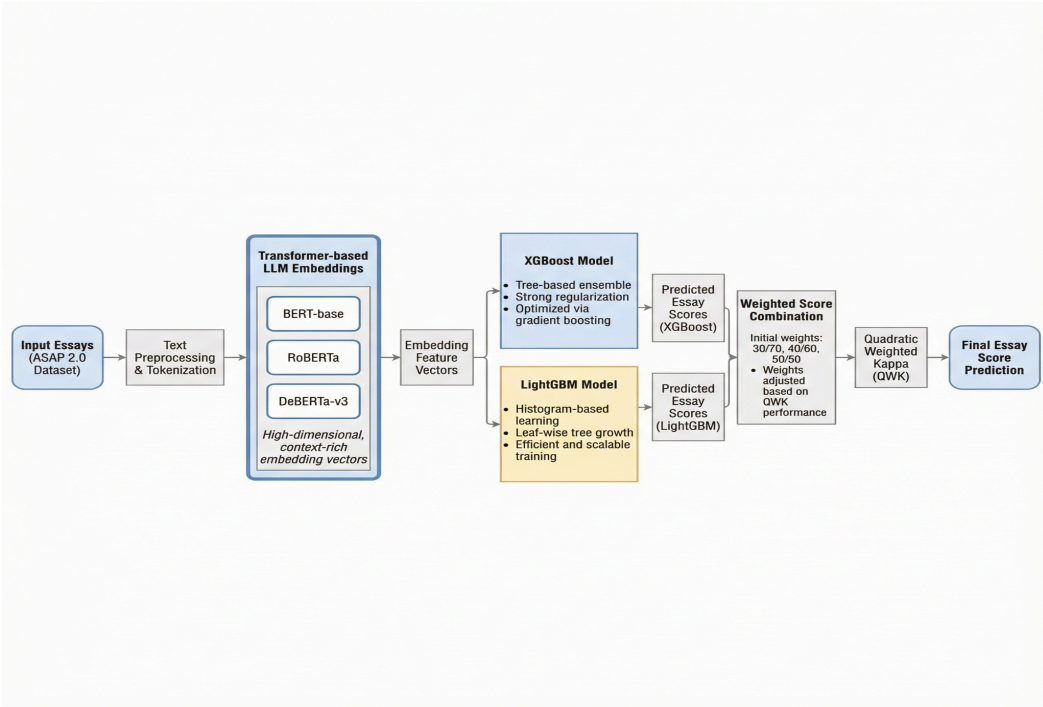

Figure 5: Tree-based Ensemble with Transformer Embeddings

One proposed approach is a hybrid framework that uses Language Model embeddings as input features and employs advanced gradient boosting algorithms to predict essay scores. This LM-embedding-based supervised learning approach first uses LM word-embeddings as basic linguistic features and feeds them into traditional machine learning algorithms. In this study, pretrained transformer-based LMs, including DeBERTa-v3, RoBERTa, and BERT-base, were adopted to extract high-dimensional, context-rich embeddings for each essay in the ASAP 2.0 dataset. Specifically, embeddings extracted from the models are used as high-dimensional feature vectors, which are then fed into two widely adopted gradient boosting methods: XGBoost and LightGBM. XGBoost (Extreme Gradient Boosting) is a scalable, tree-based ensemble learning algorithm known for its high predictive accuracy and strong regularization mechanisms, making it robust to overfitting in structured data tasks. It sequentially builds an ensemble of decision trees, optimizing a differentiable loss function using gradient descent and incorporating shrinkage and column subsampling to enhance generalization (Chen & Guestrin, 2016). LightGBM (Light Gradient Boosting Machine) focuses on computational efficiency and scalability. It employs histogram-based algorithms, leaf-wise tree growth, and Gradient-based One-Side Sampling (GOSS) to reduce memory usage and accelerate training while maintaining high accuracy (Ke et al., 2017). The embeddings extracted from LLM are implemented as input features. Both XGBoost and LightGBM are supervised learning models. They are trained separately under the same input embeddings. The predictions from both models are merged through a weight-based approach. Various sets of weights, including (30/70, 40/60, 50/50), are initially implemented and readjusted based on QWK score performance.

**Evaluation Criteria**

The performance of the automated scoring system is evaluated using the Quadratic Weighted Kappa metric, a robust statistic for measuring agreement between two sets of ordinal ratings (Cohen, 1968). QWK is particularly well-suited to automated scoring tasks, where predictions are compared against human-assigned scores that follow an ordered rubric. Unlike accuracy or mean squared error, which treat all errors equally, QWK incorporates a weighting scheme that penalizes large discrepancies more heavily than minor differences. This is essential in essay scoring, where predicting a score of 5 instead of 6 is far less severe than predicting 0 instead of 6.

$$w_{i,j} = \frac{(i-j)^2}{(N-1)^2} \qquad (1)$$

Mathematically, the weight between two score categories *i* and *j* is defined as Equation (1), where *N* is the total number of score levels. These weights are applied to both the observed and expected agreement matrices, O and E, which represent the distribution of model predictions and the agreement expected by chance, respectively. The QWK statistic is then computed as seen in Equation(2). A QWK score of 1.0 indicates perfect agreement, while a score of 0.0 reflects performance equivalent to random guessing, and negative values imply worse-than-chance predictions.

$$\kappa = 1 - \frac{\sum w_{i,j} O_{i,j}}{\sum w_{i,j} E_{i,j}} \qquad (2)$$

In the context of automated essay scoring (AES), QWK serves as the standard metric because it captures the nuanced ordinal relationships between score levels and offers a realistic measure of how well a model replicates human judgment. Human raters themselves often exhibit slight variability in their scoring, and QWK reflects this variability while emphasizing consistency and fairness. Furthermore, because essay scoring datasets are frequently imbalanced, with mid-range scores appearing more often than extreme values, QWK provides a stable evaluation framework that is less sensitive to skewed distributions than simpler metrics.

**Results**

Table 1 presents the performance of baseline models. We also add the QWK when testing with Kaggle's unpublished test set. Among SLMs, DeBERTaV3 achieved the strongest performance (QWK = 0.790), modestly outperforming BERT, RoBERTa, and DistilBERT.

*Table 1*

*The Performance of Baseline Models*

| Model | QWK Kaggle test set | QWK Our test set |
|---|---|---|
| BERT | 0.767 | 0.760 |
| RoBERTa | 0.793 | 0.760 |

| | | | |
|---|---|---|---|
| DistilBERT | | 0.756 | 0.769 |
| DeBERTaV3 | | 0.784 | 0.790 |

Table 2 summarizes ensemble results. Concatenating multiple embeddings and training an MLP improved performance to 0.815, while combining embeddings with gradient boosting further boosted QWK scores (DeBERTa + XGBoost and LightGBM = 0.812). The highest performance was achieved by a feature-based voting ensemble that integrated XGBoost, LightGBM, and a neural network, yielding 0.822 on our test set and 0.840 on Kaggle's hidden test set. These results demonstrate that ensemble approaches consistently outperform single-model baselines.

*Table 2*

*The Performance of Ensemble Models and Feature-based Model*

| Model | Input Model | Model | QWK Kaggle Test Set | QWK Our Test Set |
|---|---|---|---|---|
| Model 1 | Concatenating SLMs embeddings | | 0.807 | 0.815 |
| Model 2 | RoBERTa embeddings | XGBoost/Light GBM | 0.754 | 0.783 |
| Model 3 | DeBERTa embeddings | XGBoost/Light GBM | 0.776 | 0.812 |
| Model 4 | BERT embeddings | XGBoost/Light GBM | 0.735 | 0.781 |
| Model 5 | Linguistic features | Majority voting ensembles | 0.840 | 0.822 |

Table 3 reports the performance of top-ranked submissions on the Kaggle public leaderboard. All scores are based on the public test set, which is derived from the same dataset used throughout this study and therefore provides a consistent basis for comparison. As shown, leading models achieve closely clustered QWK scores, ranging from 0.827 to 0.833, indicating strong convergence among high-performing approaches. The top entry attains a QWK of 0.833, followed closely by several competitive solutions with marginal differences in performance.

*Table 3*

*The Performance of Leaderboard*

| Model | Team name | QWK Kaggle test set |
|---|---|---|
| Model 1 | HO | 0.833 |
| Model 2 | yao | 0.831 |
| Model 3 | GPU From onethingai.com | 0.828 |
| Model 4 | yukiZ | 0.828 |
| Model 5 | Chris Deotte | 0.827 |

## Summary and Discussion

This study presented empirical evidence to demonstrate that encoder-based language models did not perform well compared with ensemble models. Truncation of token length is expected to impair the scoring accuracy caused by loss of information of the original long essay. In addition, we fine-tuned the Flan-T5 and Gemma-2 models using Low-Rank Adaptation (LoRA), a widely adopted parameter-efficient fine-tuning (PEFT) method. Although PEFT enabled effective adaptation of these language models, their performance did not surpass that of the BERT-based baselines, suggesting that more extensive parameter tuning or alternative adaptation strategies may be required to fully exploit their capacity for this task. In contrast, despite its relatively simple architecture, the proposed ensemble model achieved superior performance by integrating embeddings from multiple LMs, highlighting the benefit of complementary representations. Notably, the resulting performance is comparable to the top scores reported on the Kaggle public leaderboard (see Table 3). This alignment indicates that our approach is competitive with state-of-the-art solutions in the broader competition and demonstrates that carefully designed ensemble strategies can match leaderboard performance without relying on heavy-weight end-to-end fine-tuning of large models.

In conclusion, our results suggest that: (1) single transformer models like BERT and RoBERTa offer less accuracy on long essay scoring; (2) parameter-efficiently tuned generative models (PEFT) without working (Flan-T5, Gemma2) did not outperform baseline SLMs, indicating that parameter-efficient tuning alone did not work well on this task; (3) long-context LMs (Longformer, BigBird) did not excel than shorter-context ones, confirming previous research that longer context windows do not yield improved AES performance; and (4) ensemble-based methods that complement embeddings with gradient boosting classifiers consistently outperformed single-model baselines as well as long-context SLMs. Notably, these results support our core contribution: the ensemble-of-embeddings approach offers a strong, lightweight alternative to long-essay scoring. Using pre-trained embeddings with lightweight learners, high agreement with human scores is achievable without expensive, large-scale fine-tuning.

Several limitations of the current study remain, including highly imbalanced score distributions, restriction to English-only essays, and limited generalizability to other assessment contexts and domains. In addition, the fixed maximum input length of encoder-based models constrains their ability to fully capture long or highly detailed responses, potentially leading to information loss for long essays. Future studies should therefore explore data augmentation either using LLMs like GPT to create more essays representing minority score categories (Jiao et al., 2024) or generating scoring rationales for data augmentation (Jiao et al., 2025). Further studies can compare ensemble strategies across more diverse corpora and languages, systematically investigate fairness and bias considerations across demographic and linguistic subgroups. In addition, some common methods such as chunking and sliding can be explored for long context input.

Beyond these directions, a promising line of recent work explores the use of text summarization by LLM like GPT as a preprocessing step for automated essay scoring (Hua, Jiao, & Wang, 2025), which demonstrated that combining a LLM-generated summary with the original essay text, rather than replacing it can effectively constrain the total input length within the maximum token limits of encoder-based language models such as BERT and RoBERTa, while preserving global discourse-level information. Empirical results show that this hybrid input strategy leads to evident performance gains compared with using truncated original text alone, suggesting that summarization can mitigate length constraints without sacrificing long essay fidelity. Incorporating such text summarization and other linguistic features into ensemble pipelines represents a promising future direction for improving accuracy and scalability in automated scoring of long essays.

Appendix A. Hyperparameter Settings for SLMs.

| Category | Hyperparameter | Value |
|---|---|---|
| **Model Architecture** | SLMs | BERT, RoBERTa, DistilBERT, DeBERTa |
| | Classification head | Linear classifier with sigmoid activation |
| | Output formulation | Ordinal multi-label (cumulative thresholds) |
| **Input Processing** | Maximum sequence length | 512 tokens |
| | Tokenization | Model-specific pretrained tokenizer |
| **Training Setup** | Epochs | 3 |
| | Batch size | 32 |
| | Random seed | 42 |
| **Optimizer** | Optimizer | Adam |
| | Base learning rate | $5 \times 10^{-6}$ |
| **Learning Rate Schedule** | Scheduler type | Cosine decay with warm-up |
| | Warm-up epochs | 2 |
| | Maximum learning rate | $9.6 \times 10^{-5}$ |
| | Minimum learning rate | $3 \times 10^{-6}$ |
| **Loss Function** | Loss | Binary cross-entropy |
| **Evaluation & Selection** | Validation metric | Weighted Quadratic Kappa (QWK) |
| | Model selection | Best validation QWK |
| **Data Pipeline** | Dataset caching | Enabled |
| | Prefetching | Enabled |
| | Drop remainder | Enabled (training and validation) |
| **Checkpointing** | Saved weights | Best model only |